\documentclass{article}
\usepackage{spconf,amsmath,graphicx}

\usepackage{times}  
\usepackage{helvet} 
\usepackage{courier}  
\usepackage[hyphens]{url}  
\usepackage{graphicx} 
\usepackage{amsmath,tabu}
\usepackage{hyperref}
\usepackage[thinlines]{easytable}
\usepackage{makecell}
\usepackage{graphicx}  
\usepackage{amssymb}
\usepackage{graphicx,multirow}
\usepackage{makecell}
\usepackage{algorithm,algorithmicx,algpseudocode}

\usepackage{booktabs}       
\usepackage{amsmath}
\usepackage{graphics}
\usepackage{xcolor}
\usepackage{tikz}

\title{Self-supervised learning for few-shot image classification}
\name{Da Chen, Yuefeng Chen, Yuhong Li, Feng Mao, Yuan He$^{\dagger}$, Hui Xue \thanks{$\dagger$ Corresponding Author.}}
\address{Alibaba Group, China \\ \{chen.cd, yuefeng.chenyf, daniel.lyh, maofeng.mf, heyuan.hy, hui.xueh@alibaba-inc.com\}}

\begin{document}
\ninept
\maketitle
\begin{abstract}
Few-shot image classification aims to classify unseen classes with limited labelled samples. Recent works benefit from the meta-learning process with episodic tasks and can fast adapt to class from training to testing. Due to the limited number of samples for each task, the initial embedding network for meta-learning becomes an essential component and can largely affect the performance in practice.
%
%
To this end, most of the existing methods highly rely on the efficient embedding network. Due to the limited labelled data, the scale of embedding network is constrained under a supervised learning(SL) manner which becomes a bottleneck of the few-shot learning methods.
In this paper, we proposed to train a more generalized embedding network with self-supervised learning (SSL) which can provide robust representation for downstream tasks by learning from the data itself. 
We evaluate our work by extensive comparisons with previous baseline methods on two few-shot classification datasets ({\em i.e.,} MiniImageNet and CUB) and achieve better performance over baselines. Tests on four datasets in cross-domain few-shot learning classification show that the proposed method achieves state-of-the-art results and further prove the robustness of the proposed model.
%
%
Our code is available at \hyperref[https://github.com/phecy/SSL-FEW-SHOT.]{https://github.com/phecy/SSL-FEW-SHOT.}
\end{abstract}
\section{Introduction}
\label{sec:intro}

Recent advances in deep learning techniques have made significant progress in many areas. The main reason for such success is the ability to train a deep model that can retain profound knowledge from large scale labelled dataset. This is somehow against human learning behaviour - one can easily classify objects from just a few examples with limited prior knowledge. How to computationally model such behaviour motivates the recent researches in few-shot learning, where the focus is on how to adapt the model to new data or tasks with a restricted number of instances. 

One popular solution for few-shot classification is to apply a fine-tuning process on existing embedding network to adapt new classes. The main challenge here is that the fine-tuning could easily lead to overfitting, as only a few samples(1-shot or 5-shot) for each class are available. 
One recent proposed solution for few-shot classification is a meta-learning process, in which the dataset is divided into subsets for different meta tasks to learn how to adapt the model according to the task change. These methods highly rely on an effective pre-train embedding network.
%
%
%

Current methods~\cite{jiang2018learning_caml,oreshkin2018tadam,qiao2018few_shot_activations,rusu2018LEO} with good performance mostly apply a ResNet12~\cite{he2016residual} or a wide ResNet~\cite{zagoruyko2016wide_residul_WRN} as the embedding network and surpass the methods~\cite{bauer2017discriminative_k_shot,chen2019aug_few_shot} with deeper network. We argue that the abandon of large network is mainly because all these methods are trained in a supervised way with limited labelled samples.
In this paper, we propose to apply a much larger embedding network with self-supervised learning (SSL) to incorporate with episodic task based meta-learning. According to the evaluation presented in Section~\ref{sec:result}, the proposed method can significantly improve few-shot image classification performance over baseline methods in two common datasets. As a remark, under the the same experiment setting, the proposed method improves 1-shot and 5-shot tasks by nearly \textbf{3\%} and \textbf{4\%} on MiniImageNet, by nearly \textbf{9\%} and \textbf{3\%} on CUB. Moreover, the proposed method can gain the improvement of \textbf{15\%}, \textbf{13\%} and \textbf{15\%}, \textbf{8\%} in two tasks on MiniImageNet and CUB dataset by pretraining using more unlabeled data. We also observe that the proposed model can be robustly transferred to other datasets under a recently proposed cross-domain few-shot learning scenario~\cite{guo2020broader_cross_domain_fewshot} and achieve the state-of-the-art result(\textbf{69.69\%} vs. 68.14\%).

\begin{figure*}[h]
\begin{center}
\includegraphics[width=0.85\linewidth]{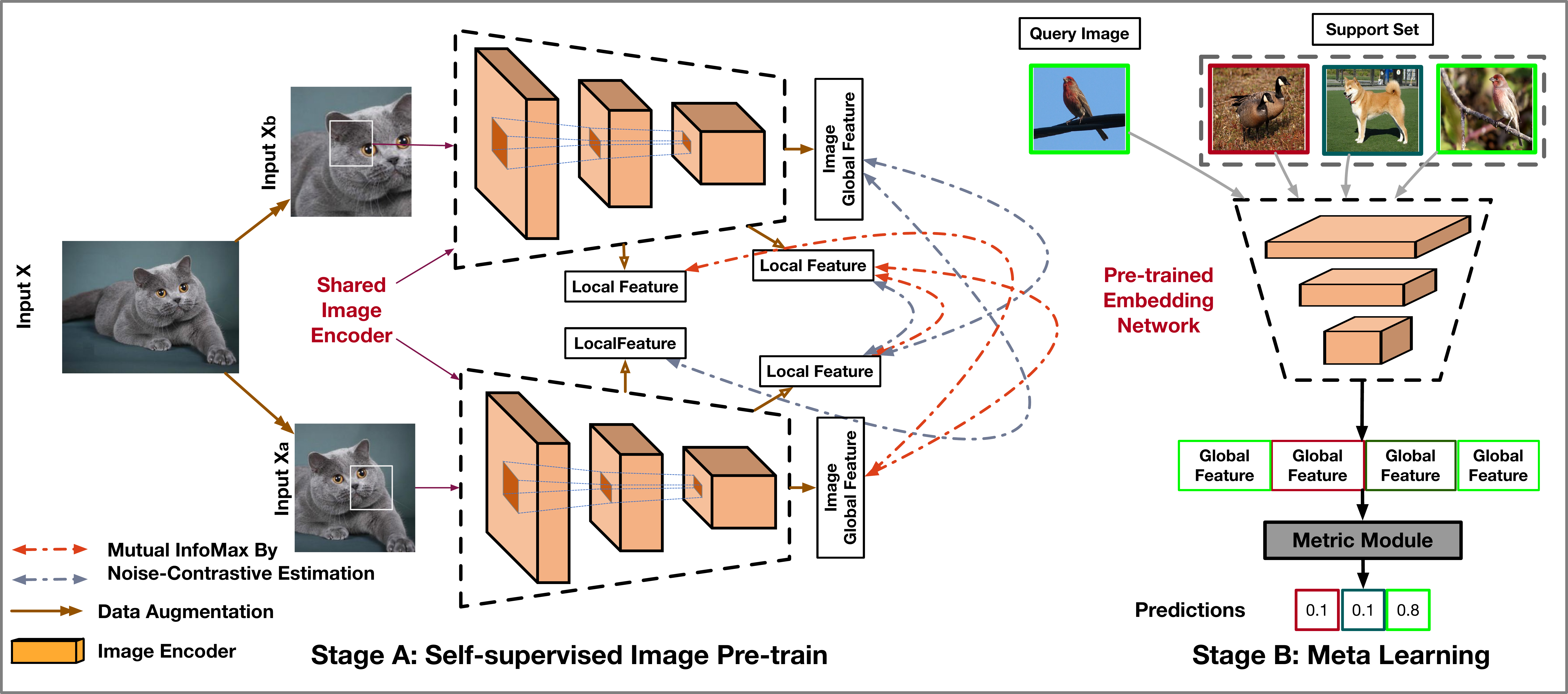}
\end{center}
\vspace*{-6mm}
\caption{The overall architecture of our approach. LEFT: Train embedding network by Self-Supervised learning. The pretext task is designed to maximize the mutual information between two views $\left<x_a, x_b\right>$ generated from the same image $x$ by data augmentation. Right: Meta-learning with an episodic task(3-way, 1-shot example). For each task, the training samples and query samples are encoded by the embedding network. Query sample embeddings are compared with the centroid of training sample embeddings and make a further prediction.}
\vspace*{-3mm}
\label{fig:archi}  
\end{figure*}

\section{Related Work}
\label{sec:rel}

\noindent\textbf{Few-shot learning} as an active research topic has been extensively studied. In this paper, we will primarily review recent deep-learning based approaches that are more relevant to our work. A number of works aim to improve the robustness of the training process. Zhao~\textit{et al.}~\cite{zhao2018msplit_few_shot} split the features to three orthogonal parts to improve the classification performance for few-short learning, allowing simultaneous feature selection and dense estimation. Chen~\textit{et al.}~\cite{chen2019aug_few_shot} propose a Self-Jig algorithm to augment the input data in few-shot learning by synthesizing new images that are either labelled or unlabeled. 

A popular strategy for few-shot learning is through meta-learning (also called learning-to-learn) with multi-auxiliary tasks~\cite{nips2016_vinyals2016matching_net,finn2017maml,sung2018relation_net,qiao2018few_shot_activations}. The key is how to robustly accelerate the learning progress of the network without suffering from over-fitting  with limited training data. Finn~\textit{et al.} propose MAML~\cite{finn2017maml} to search the best initial weights through gradient descent for network training, making the fine-tuning easier. REPTILE~\cite{nichol2018reptile} simplifies the complex computation of MAML by incorporating an $L_2$ loss, but still performs in high dimension space. To reduce the complexity, Rusu~\textit{et al.} propose a network called LEO~\cite{rusu2018LEO} to learn a low dimension latent embedding of the model.
Another stream of meta-learning based approaches~\cite{snell2017prototypical_net, oreshkin2018tadam,nips2016_vinyals2016matching_net,sung2018relation_net} attempt to learn a deep embedding model that can effectively project the input samples to a specific feature space. Then the samples can be classified by the nearest neighbour (NN) criterion using a pre-defined distance function.
 Deeper embedding backbone has also been tried. Chen~\textit{et al.}~\cite{chen2019aug_few_shot} propose a data augmentation method to cope with the over-fitting issue with deeper backbone embedding network. In~\cite{hariharan2017low_hallucinating_features}, Hariharan~\textit{et al.} indicate similar observation, as claimed in this paper, that with a deeper backbone(ResNet-50) as embedding is not only costly but also not effective and proposed a novel representation regularization techniques.
%

\noindent\textbf{Self-supervised learning(SSL)} aims to learn robust representations from the data itself without class labels. The main challenge here is how to  design the pretext tasks that are complex enough to exploit high-level compact semantic visual representations that are useful for solving downstream tasks. This is consistent with the mission of the pre-trained embedding network in few-shot learning. The work of~\cite{KolesnikovZB19} revisited some state-of-the-art methods based on various classification based pretext tasks({\em e.g.,} Rotation, Exemplar, RelPatchLoc, Jigsaw). Recently, several methods have been proposed to combine SSL with few-shot learning. Gidaris~\textit{et al.}~\cite{gidaris2019boosting_fewshot_ssl} proposed to apply a combination of supervised loss and self-supervised loss to pre-train the embedding network. A study about embedding network in few-shot learning~\cite{tian2020rethinking_fewshot_embedding} indicates that self-supervised learning based embedding network can achieve similar results as supervised learning. All these methods indicate that with a simple network such as ResNet-12, ResNet-18 can outperform ResNet-50. Different from these work, the proposed method does not rely on any supervised learning during the pre-training phase and prove that large scale embedding network can be applied in few-shot learning and achieve better results.
%


\section{Method}
\label{sec:method}

Few-shot learning is a challenging problem as it has only limited data for training and needs to verify the performance on the data for unseen classes. An effective solution for few-shot learning classification problem is to apply a meta-learning scheme on top of a pre-trained embedding network. Most of the current methods are mainly focusing on the second stage~{\em i.e., meta-learning stage}. In this work, we follow this two stages paradigm but utilize self-supervised learning to train a large embedding network as our strong base in stage one.

\subsection{Self-supervised learning stage}
\label{subsec:self_supervised_learning}
Our goal is to learn representations that enhance the feature's generalization.  In our approach, we use Augmented Multiscale Deep InfoMax (AMDIM)~\cite{AMDIM} as our self-supervised model. The pretext task is designed to maximize the mutual information between features extracted from multiple views of a shared context.  

The mutual information (MI) measures the shared information between two random variables $X$ and $Y$  which is defined as the Kullback–Leibler (KL) divergence between the joint and the product of the marginals.
\begin{equation}
\begin{split}
   I( X,Y ) = D_{KL}\left( p(x,y) || p(x)p(y)\right) \\
   = \sum \sum p(x,y) \log \frac{p(x|y)}  {p(x)} 
\end{split}
\end{equation}
where $P(x,y)$ is the joint distribution and  $P(x)$ and $P(y)$ are the marginal distributions of $X$ and $Y$. Estimating MI is challenging as we just have samples but not direct access to the underlying distribution. ~\cite{oord2018representation} proved that we can maximize a lower bound on mutual information by minimizing the Noise Contrastive Estimation (NCE) loss based on negative sampling.


 The core is to maximize mutual information between global features and local features from two views $(x_a, x_b)$ of the same image. Specifically, maximize mutual information between $\left<f_g(x_a), f_5(x_b)\right>$, $\left<f_g(x_a), f_7(x_b)\right>$ and $\left<f_5(x_a), f_5(x_b)\right>$.  
Where $f_g$ is the global feature, $f_{5}$ is encoder's  $5 \times 5$ local feature map as well as $f_7$  as the encoder's $7 \times 7$ feature map. For example, the NCE loss between $f_g(x_a)$ and $f_5 ( x_b )$ is defined as below:
\begin{equation}
\begin{split}
& \mathcal{L}_{ssl}\left(f_g(x_a) , f_5 ( x_b )\right) = \\
&  -\log  \frac{ \exp\{\phi(f_g(x_a), {{f_5}( x_b )}) \} }{\sum_{\widetilde{x_b} \in {\mathcal{N}_{x} \cup	x_b}} \exp\{ \phi(f_g(x_a), {f_5}( \widetilde{x_b} )) \} }
\end{split}
\end{equation}
$\mathcal{N}_x$ are the negative samples of image $x$, $\phi$ is the distance metric function. At last, the overall loss between $x_a$ and $x_b$ is as follows:
\begin{equation}
\begin{split}
\mathcal{L}_{ssl}( x_a, x_b) =  \mathcal{L}_{ssl}\left(f_g(x_a) , f_5 ( x_b )\right) \ + \\  \mathcal{L}_{ssl}\left(f_g(x_a) , f_7 ( x_b )\right) +  \mathcal{L}_{ssl}\left(f_5(x_a) , f_5 ( x_b )\right)
\end{split}
\end{equation}
For more details, please refer to~\cite{AMDIM}.
\subsection{Meta-learning stage}
\label{subsec:meta_learing_general}

Given a pre-trained embedding network from stage one, meta-learning is applied to further fine-tune the model with an episodic manner. A few-shot $K$-way image classification task can be illustrated as a $K$-way $C$-shot problem~{\em i.e.,} given $C$ labelled samples for each unseen class, the model should fast adapt to them to classify novel classes. The entire training set can be presented by $D=\{(x_1,y_1),\dots,(x_N,y_N)\}$ where $N$ is the total number of classes in $D$, $x$ is the training sample with label $y$. For a specific $K$-way $C$-shot meta task $T$, $V = \{y_i|i=1,\dots,K\}$ denotes the class labels randomly chosen from $D$. Training samples from these classes are randomly chosen to form a support set and a query set: (\textbf{a}) the support set for task $T$ is denoted by $S$, which contains $C\times K$ samples  ($K$-way $C$-shot); (\textbf{b}) the query set is $Q$ where $n$ is the number of samples selected for meta testing.

In this paper, during the meta-learning stage, the proposed model is trained to learn an embedding function to map all input samples from same class to a mean vector $c$ in a description space as a class descriptor for each class~\cite{snell2017prototypical_net}. For class $k$, it is represented by the centroid of embedding features of training samples and can be obtained as:

\begin{equation}
\label{equ:pnet_ck}
    c_k = \frac{1}{\left | S_k \right |}\sum_{(x_i,y_i)\in S} f (x_i),
\end{equation}
where $f(x_i)$ is the embedding function initialized by stage one, $S_k$ is the training samples labelled with class $k$.

As a metric learning based method, we employ a distance function $d$ and produce a distribution over all classes given a query sample $q$ from the query set $Q$:

\begin{equation}
\label{equ:loss1}
\begin{split}
p(y = k|q) = \frac{\exp(-d(f(q),c_k))}{\sum_{k'}\exp(-d(f(q),c_{k'}))}
\end{split}
\end{equation}

In this paper, Euclidean distance is chosen as distance function $d$. As shown in Eq.~\ref{equ:loss1}, the distribution is based on a softmax over the distance between the embedding of the samples (in the query set) and the class descriptors. The loss in the meta-learning stage can then read:

\begin{equation}
\label{equ:loss2}
 \mathcal{L}_{meta} = d(f(q),c_k) + \log\sum_{k'}d(f(q),c_{k'})
\end{equation}

In a short conclusion, the proposed method first applies an SSL way to pre-train a large scale embedding network in stage one, followed by a detailed fine-tuning in stage two with a meta-learning scheme. 

\section{Experimental Results}
\label{sec:result}

In this section, we first introduce the dataset and training process used in our evaluation, then show quantitative comparisons against other baseline methods, finally we conduct a detailed study to validate the transferability of our approach under a cross-domain few-shot learning evaluation set-up proposed in~\cite{guo2020broader_cross_domain_fewshot}.

\subsection{Datasets}
\label{subsec:dataset}

MiniImageNet dataset~\cite{nips2016_vinyals2016matching_net}, is a subset of ImageNet which is a standard benchmark to evaluate the performance of few-shot learning methods. It contains 60,000 images from 100 classes, and each class has 600 images. We follow the data split strategy in~\cite{ravi2016optimization} to sample images of 64 classes for training, 16 classes for validation, 20 classes for the test.

CUB-200-2011(CUB) dataset, proposed in~\cite{WahCUB_200_2011}, is a dataset for fine-grained classification. It contains 200 classes of birds with 11788 images in total. For evaluation, we follow the split in~\cite{hilliard2018few_conditional_embedding_maco}. 200 species of birds are randomly split to 100 classes for training, 50 classes for validation, and 50 classes for the test.

For cross-domain few-shot learning, 4 datasets are proposed to test suggested by~\cite{guo2020broader_cross_domain_fewshot},~{\em i.e.,} 1) CropDiseases~\cite{mohanty2016using_plant}, a plant diseases dataset, 2) EuroSAT~\cite{helber2019eurosat}, a dataset for satellite images,  2) ISIC~\cite{codella2019ISIC_2} a medical skin image dataset, 4) ChestX~\cite{wang2017chestx}, a dataset for X-ray chest images. The similarity comparing to MiniImageNet is decreasing across these datasets.

\subsection{Training Details}
\label{subsec:trainingg_details}
Several recent works show that a typical training process can include a pre-trained network~\cite{qiao2018few_shot_activations,rusu2018LEO} or employ co-training~\cite{oreshkin2018tadam} for feature embedding. This can significantly improve the classification accuracy. In this paper, we adopt the AMDIM ~\cite{AMDIM} SSL training framework to pre-train the feature embedding network. AmdimNet(ndf=192, ndepth=8, nrkhs=1536) is used for all datasets and the embedding dimension is 1536. Adam is chosen as the optimizer with a learning rate of $0.0002$. We use $128 \times 128$ as the input resolution of unlabelled images among these datasets for self-supervised training. During meta-learning stage, image size is down to $84 \times 84$ following previous methods. For MiniImageNet dataset, 3 embedding models are trained. \textbf{Mini80-SSL} is self-supervised trained from  48,000 images (80 classes training and validation ) without labels. \textbf{Mini80-SL} is supervised training using same AmdimNet by cross-entropy loss with labels. \textbf{Image900-SSL} is SSL trained from all images from ImageNet1K except MiniImageNet. For CUB dataset,  \textbf{CUB150-SSL} is trained by SSL from 150 classes (training and validation). \textbf{CUB150-SL} is the supervised trained model.~\textbf{Image1K-SSL} is SSL trained from  all images from ImageNet1K without any label. 
For cross-domain test, \textbf{Mini80-SSL} is applied as embedding network across all tests in four datasets during training.

\begin{table}[]
\centering
 \scalebox{0.8}{
\begin{tabular}{l|c c c}
\Xhline{3\arrayrulewidth}
\textbf{Baselines}  & \textbf{Embedding Net} & \textbf{1-Shot 5-Way} & \textbf{5-Shot 5-Way} \\ 
\Xhline{2\arrayrulewidth}
MatchingNet~\cite{nips2016_vinyals2016matching_net}  & 4 Conv   & 43.56 $\pm$ 0.84\%   & 55.31 $\pm$ 0.73\%               \\ 
MAML~\cite{finn2017maml}      & 4 Conv            & 48.70 $\pm$ 1.84\%   & 63.11 $\pm$ 0.92\%             \\ 
RelationNet~\cite{sung2018relation_net} & 4 Conv           & 50.44 $\pm$ 0.82\%   & 65.32 $\pm$ 0.70\%            \\ 
REPTILE~\cite{nichol2018reptile}      &   4 Conv       & 49.97 $\pm$ 0.32\%   & 65.99 $\pm$ 0.58\%                  \\ 
ProtoNet~\cite{snell2017prototypical_net} &     4 Conv         & 49.42 $\pm$ 0.78\%   & 68.20 $\pm$ 0.66\%             \\ 
Baseline*~\cite{chen2018a}  &       4 Conv        & 41.08 $\pm$ 0.70\%         & 54.50\% $\pm$ 0.66              \\ 
Spot\&learn~\cite{chu_cvpr2019_spot_and_learn}  &       4 Conv        & 51.03 $\pm$ 0.78\%         & 67.96\% $\pm$ 0.71                 \\ 
DN4~\cite{li_cvpr2019_revisiting}  &       4 Conv        & 51.24 $\pm$ 0.74\%         & 71.02\% $\pm$ 0.64                  \\ \hline

SNAIL~\cite{mishra2017snail}    & ResNet12
& 55.71 $\pm$ 0.99\%   & 68.88 $\pm$ 0.92\%                \\ 
ProtoNet$^{+}$~\cite{snell2017prototypical_net} & ResNet12
& 56.50 $\pm$ 0.40\% & 74.2 $\pm$ 0.20\%    \\


MTL~\cite{sun2019meta_mtl}        & ResNet12
& 61.20 $\pm$ 1.8\%   & 75.50 $\pm$ 0.8\%                \\
DN4~\cite{li_cvpr2019_revisiting}        & ResNet12
& 54.37 $\pm$ 0.36\%   & 74.44 $\pm$ 0.29\%               \\
TADAM~\cite{oreshkin2018tadam}      &    ResNet12
& 58.50\%         & 76.70\%                 \\

Qiao-WRN~\cite{qiao2018few_shot_activations}   &    Wide-ResNet28
& 59.60 $\pm$ 0.41\%   & 73.74 $\pm$ 0.19\%            \\ 
LEO~\cite{rusu2018LEO}        & Wide-ResNet28
& 61.76 $\pm$ 0.08\%   & 77.59 $\pm$ 0.12\%                \\
Dis. k-shot~\cite{bauer2017discriminative_k_shot} & ResNet34  & 56.30 $\pm$ 0.40\%   & 73.90 $\pm$ 0.30\%       \\ 
Self-Jig(SVM)~\cite{chen2019aug_few_shot}     &   ResNet50
& 58.80 $\pm$ 1.36\%   & 76.71 $\pm$ 0.72\%             \\ 
FEAT~\cite{ye2020feat}     &   ResNet50
& 53.8\%   & 76.0\%             \\
\Xhline{3\arrayrulewidth}
\textbf{Ours\_Mini80\_SL}      &  AmdimNet   & 43.92 $\pm$ 0.19\%  & 67.13 $\pm$ 0.16\%     \\
\textbf{Ours\_Mini80\_SSL$^{-}$}      &  AmdimNet   & 46.13 $\pm$ 0.17\%  & 70.14 $\pm$ 0.15\%     \\
\textbf{Ours\_Mini80\_SSL}      &  AmdimNet   & \textbf{64.03 $\pm$ 0.20\%}  & \textbf{81.15 $\pm$ 0.14\%}     \\
\textbf{Ours\_Image900\_SSL}      &  AmdimNet   & \textbf{76.82 $\pm$ 0.19\%}  & \textbf{90.98 $\pm$ 0.10\%}     \\
\Xhline{3\arrayrulewidth}
\end{tabular}
}
\caption{Few-shot classification accuracy results on \textit{MiniImageNet}. $'-'$ indicates result without meta-learning.
}
\vspace{-4mm}
\label{tab:Mini_result}
\end{table}

\subsection{Evaluation results}
\label{subsec:Quantitative_comparison}
\subsubsection{Standard few-shot learning evaluation}

For MiniImageNet and CUB, we evaluate our method in two common few-shot learning tasks~{\em i.e.,} 1-shot 5-way task and 5-shot 5-way task against baseline methods with different embedding networks including classical ones~\cite{snell2017prototypical_net,nips2016_vinyals2016matching_net,finn2017maml} and recently proposed methods~\cite{rusu2018LEO,oreshkin2018tadam,ye2020feat}. For CUB dataset, we follow the work~\cite{chen2018a} to evaluate the robustness of the proposed framework with 7 other alternatives on this fine-grained dataset.

As detailed in Table~\ref{tab:Mini_result}, the proposed method outperforms all baselines in the tested tasks. In 1-shot 5-way test, our approach achieves $7.53\%$ and $2.27\%$ improvement over ProtoNet$^{+}$~\cite{snell2017prototypical_net} and LEO~\cite{rusu2018LEO} respectively. The former is an amended variant of ProtoNet using pre-trained Resnet as embedding network and has the same meta-learning stage with the proposed method. In the experience for 5-Shot 5-Way, we observe a similar improvement. Furthermore, we observe that the performance of our proposed method significantly increases when receiving more images/classes as input for pre-train. With more unlabeled samples(Image900\_SSL), the model can achieve average 16\% improvement over baselines while applies the same amount of labelled data.  

Table~\ref{tab:CUB_result} illustrates our experiment on CUB dataset. Our proposed method yields the highest accuracy from all trials. In the 1-shot 5-way test, we have $71.85\%$ gaining a margin of $20.54\%$ increment to the classic ProtoNet~\cite{snell2017prototypical_net}. The improvement is more significant for the 5-shot 5-way test. Our proposed method results is $84.29\%$ which introduces $2.39\%$ improvement to DN4-Da~\cite{li_cvpr2019_revisiting}. Comparing to Baseline++~\cite{chen2018a}, our method shows a significant improvement, i.e., $11.32\%$ and $4.95\%$ in both tests.

\begin{table}[]
\centering
\scalebox{0.8}{
\begin{tabular}{l|c c c}
\Xhline{3\arrayrulewidth}
\textbf{Baselines}      & Embedding Net & \textbf{1-Shot 5-Way} & \textbf{5-Shot 5-Way} \\ 
\Xhline{2\arrayrulewidth}
MatchingNet~\cite{nips2016_vinyals2016matching_net} & 4 Conv & 61.16 $\pm$ 0.89   & 72.86 $\pm$ 0.70   \\ 
MAML~\cite{finn2017maml}    &  4 Conv  & 55.92 $\pm$ 0.95\%   & 72.09 $\pm$ 0.76\%   \\ 
ProtoNet~\cite{snell2017prototypical_net} & 4 Conv  & 51.31 $\pm$ 0.91\%   & 70.77 $\pm$ 0.69\%   \\ 
MACO~\cite{hilliard2018few_conditional_embedding_maco}   &  4 Conv   & 60.76\%        & 74.96\%        \\ 
RelationNet~\cite{sung2018relation_net} & 4 Conv & 62.45 $\pm$ 0.98\%   & 76.11 $\pm$ 0.69\%   \\ 
Baseline++~\cite{chen2018a} & 4 Conv & 60.53 $\pm$ 0.83\%   & 79.34 $\pm$ 0.61\%   \\ 
DN4-DA~\cite{li_cvpr2019_revisiting} & 4 Conv & 53.15 $\pm$ 0.84\%   & 81.90 $\pm$ 0.60\%   \\ 
\Xhline{3\arrayrulewidth}
\textbf{Ours\_CUB150\_SL}    & AmdimNet   & 45.10  $\pm$ 0.21\%  & 74.59 $\pm$ 0.16\%   \\
\textbf{Ours\_CUB150\_SSL$^{-}$}    & AmdimNet   & 40.83  $\pm$ 0.16\%  & 65.27 $\pm$ 0.18\%   \\
\textbf{Ours\_CUB150\_SSL}    & AmdimNet   & \textbf{71.85  $\pm$ 0.22\%}  & \textbf{84.29 $\pm$ 0.15\%}   \\
\textbf{Ours\_Image1K\_SSL}    & AmdimNet   & \textbf{77.09  $\pm$ 0.21\%}  & \textbf{89.18 $\pm$ 0.13\%}   \\
\Xhline{3\arrayrulewidth}
\end{tabular}
}
\caption{Few-shot classification accuracy results on CUB dataset~\cite{WahCUB_200_2011}. $'-'$ indicates result without meta-learning.
For each task, the best-performing method is highlighted.}
\vspace{-1mm}
\label{tab:CUB_result}
\end{table}

\subsubsection{Cross-domain few-shot Learning}

Follow the set-up in~\cite{guo2020broader_cross_domain_fewshot}, the proposed method is tested on four datasets across three tasks~{\em i.e.,} 5-way 5-shot, 5-way 20-shot 5-way 50-shot. Only Mini80\_SSL embedding network(training only with MiniImageNet 80 classes) is applied in this section. We adopt the same transductive learning set-up proposed in~\cite{guo2020broader_cross_domain_fewshot} for few-shot learning. As suggested in~\cite{guo2020broader_cross_domain_fewshot}, results across all tests on four datasets are averaged. As shown in Tab.~\ref{tab:result_tab}, after averaging the test results, the proposed method has nearly 1.5\% improvement over the state-of-the-art results which further prove the robustness of the proposed model. Noted that the state-of-the-art method obtain less accuracy in ISIC dataset, which needs more investigation in the future work.

\begin{table}[]
\centering
\scalebox{0.5}{
\begin{tabular}{@{}ccccccc@{}}
\toprule
Methods            & \multicolumn{3}{c}{ChestX}                                                                    & \multicolumn{3}{c}{ISIC}                     \\ \midrule
                   & 5-way 5-shot                 & 5-way 20-shot                 & 5-way 50-shot                 & 5-way 5-shot                 & 5-way 20-shot                 & 5-way 50-shot \\ \cmidrule(l){2-7} 
Ours\_trans         & \textbf{28.50 $\pm$ 0.40\%}           & \textbf{33.79 $\pm$ 0.48\%}            & \textbf{38.78 $\pm$ 0.64\%}            & 44.15 $\pm$ 0.52\%           & 55.63 $\pm$ 0.49\%            & 62.76 $\pm$ 0.50\%            \\
Cross~\cite{guo2020broader_cross_domain_fewshot} & 26.09 $\pm$ 0.96\%           & 31.01 $\pm$ 0.59\%            & 36.79 $\pm$ 0.53\%            & \textbf{49.68 $\pm$ 0.36\% }                       & \textbf{61.09 $\pm$ 0.44\%}                    & \textbf{67.20 $\pm$ 59\%}                      \\
\midrule
                   & \multicolumn{3}{c}{EuroSAT}                                                                  & \multicolumn{3}{c}{CropDiseases}             \\ \midrule
                   & 5-way 5-shot                 & 5-way 20-shot                 & 5-way 50-shot                 & 5-way 5-shot                 & 5-way 20-shot                 & 5-way 50-shot \\ \cmidrule(l){2-7} 
Ours\_trans   & \textbf{83.44 $\pm$ 0.61\%}           & \textbf{90.43 $\pm$ 0.52\%}            & \textbf{94.71 $\pm$ 0.47\%}            & \textbf{91.79 $\pm$ 0.48\%}           & \textbf{97.38 $\pm$ 0.65\%}            & \textbf{99.50 $\pm$ 0.63\%}      \\

Cross\cite{guo2020broader_cross_domain_fewshot} & 81.76 $\pm$ 0.48\%           & 87.97 $\pm$ 0.42\%            & 92.00 $\pm$ 0.56\%            & 90.64 $\pm$ 0.54\%                         & 95.91 $\pm$ 0.72\%                    & 97.48 $\pm$ 0.56\%                      \\
\midrule
\end{tabular}
}
\caption{Cross-domain few-shot learning tests on four datasets}
\vspace{-4mm}
\label{tab:result_tab}
\end{table}

\subsection{Ablation Study}
\label{subsec:self_com}
As shown in the quantitative evaluation, the proposed method can significantly improve the performance in few-shot classification task by including a large scale embedding network. One concern that may be raised is that if the gain of improvements of the proposed network is simply due to the increment of the network's capacity. To prove the effectiveness of the proposed method, we train the embedding network with labelled data (Mini80-SL and CUB150-SL as detailed in Section~\ref{subsec:trainingg_details}). As shown in Table~\ref{tab:Mini_result} and Table~\ref{tab:CUB_result}, it performs even worse than the methods with simple 4 Conv blocks embedding networks as such big network under supervised learning with limited data can cause overfitting problem and cannot adjust to new unseen classes during testing. However, with SSL based pre-training a more generalized embedding network can be obtained and improve the results significantly. One may also concern about the effectiveness of the meta-learning fine-tuning in the second stage. To test this, the pre-train embedding network is directly applied to the task with the nearest neighbourhood(NN) classification. As shown in the test results on both dataset, meta-learning can effectively fine-tune the embedding network and achieve remarkable improvement. 

We also include more data without labels during SSL pre-training and observe a more significant improvement of the result. As shown in Table~\ref{tab:Mini_result}, the proposed method can gain the improvement of 15\% and 13\% in two test tasks. As detailed analyzed in~\cite{chen2018a}, current few-shot learning methods can not efficiently transfer the domain of learning, {\em i.e.,} the training domain can not have a huge gap with the testing set. In this paper, a transferability test is also conducted by pre-training the embedding network on ImageNet and applied on CUB dataset. As shown in Table~\ref{tab:CUB_result}, the proposed method with ImageNet pre-trained embedding network can be efficiently transferred to CUB dataset and gain an improvement of 15\%,8\% in both test tasks. Tests on four extra datasets suggested by~\cite{guo2020broader_cross_domain_fewshot} further prove the transferability of the proposed method.

\section{Conclusion}
\label{sec:conclusion}

In this paper, we propose to utilize self-supervised learning to efficiently train a robust embedding network for few-shot image classification. The resulted embedding network is more generalized and more transferable comparing to other baselines. After fine-tuning by a meta-learning process, the performance of the proposed method can significantly outperform all baselines based on the quantitative results using two common few-shot classification datasets and a cross-domain few-shot learning set-up. The current framework can be extended in several ways in the future. For instance, one direction is to combine these two stages together and develop an end-to-end method for this task. Another direction is to investigate the effectiveness of the proposed method on another few-shot tasks such as few-shot detection,~{\em etc.}

\section{Acknowledgement}
\label{sec:Acknowledgement}
This work was supported by Alibaba Group.

{
\small
\bibliographystyle{IEEEbib}
\bibliography{ref}
}
\end{document}